\documentclass[journal]{IEEEtran}
\IEEEpubid{\makebox[\columnwidth]{979-8-3503-7903-7/24/\$31.00 \copyright2024 European Union\hfill} \hspace{\columnsep}\makebox[\columnwidth]{ }}

\usepackage[utf8]{inputenc} 
\usepackage[T1]{fontenc}    
\usepackage{hyperref}       
\usepackage{url}            
\usepackage{booktabs}       
\usepackage{amsfonts}       
\usepackage{nicefrac}       
\usepackage{microtype}      
\usepackage{xcolor}         
\usepackage{float}
\usepackage{graphicx}
\usepackage{siunitx}
\usepackage{placeins}

\usepackage{tabularx}
\usepackage{array}

\usepackage{amssymb}
\usepackage{amsfonts} 
\usepackage{amsmath}
\usepackage{amssymb}
\usepackage[caption=false, font=footnotesize]{subfig}

\title{Vision Transformers for Weakly-Supervised Microorganism Enumeration}

\author{
    \IEEEauthorblockN{Javier Ure\~{n}a Santiago\IEEEauthorrefmark{1}, Thomas Str\"{o}hle\IEEEauthorrefmark{2}\IEEEauthorrefmark{3}, Antonio Rodr\'{i}guez-S\'{a}nchez\IEEEauthorrefmark{1},  Ruth Breu\IEEEauthorrefmark{2}}\\
    \IEEEauthorblockA{\IEEEauthorrefmark{1}Intelligent and Interactive Systems, University of Innsbruck, Austria}\\
    \IEEEauthorblockA{\IEEEauthorrefmark{2}Quality Engineering, University of Innsbruck, Austria}
    \IEEEauthorblockA{\IEEEauthorrefmark{3}University of Applied Sciences Kufstein Tirol, Austria}
}

\begin{document}
\maketitle
\begin{abstract}
Microorganism enumeration is an essential task in many applications, such as assessing contamination levels or ensuring health standards when evaluating surface cleanliness. However, it's traditionally performed by human-supervised methods that often require manual counting, making it tedious and time-consuming. Previous research suggests automating this task using computer vision and machine learning methods, primarily through instance segmentation or density estimation techniques. This study conducts a comparative analysis of vision transformers (ViTs) for weakly-supervised counting in microorganism enumeration, contrasting them with traditional architectures such as ResNet and investigating ViT-based models such as TransCrowd. We trained different versions of ViTs as the architectural backbone for feature extraction using four microbiology datasets to determine potential new approaches for total microorganism enumeration in images. Results indicate that while ResNets perform better overall, ViTs performance demonstrates competent results across all datasets, opening up promising lines of research in microorganism enumeration. This comparative study contributes to the field of microbial image analysis by presenting innovative approaches to the recurring challenge of microorganism enumeration and by highlighting the capabilities of ViTs in the task of regression counting.

\end{abstract}
\section{Introduction}
Enumeration of microorganisms is crucial across diverse sectors including medicine, pharmaceutical quality control, or environmental monitoring \cite{liu2004high, riepl2011applicability, kepner1994use}. This task is relevant in microbiology, and methods have been researched and improved for decades \cite{Herbert_1990}. Traditional techniques are usually tedious, as counting manually (agar plate or hemocytometry) or the indirect estimations with turbidimetry \cite{Horwitz_1970, Absher_1973_hemocytometry, Zhou_2015_turbidimetry} depends on specialized equipment, team, and time in order to improve efficiency. Hence, there has been a pursuit of automated and efficient enumeration techniques to improve this practice \cite{zhang_comprehensive_2022}.

As a result, the efficiency of computer vision methods has improved microbial enumeration in later years. Machine learning, and later, deep learning, has made it possible to count the number of particles in images and extract various characteristic parameters of them, reducing workload and improving the accuracy of the analysis \cite{spahn_deepbacs_2021, von_chamier_democratising_2021}. Research to improve numeration through deep learning is extensive and follows two lines of research: detection-based methods and regression-based methods, like image masks estimation, or density map regression \cite{Hoorali_UNet_2020, he_deeply-supervised_CNN_density_2020}. These approaches have set the baseline and brought a new standard for efficiency all over the academic corpus.

\begin{figure}[H]
    \centering
    \includegraphics[width=\linewidth]{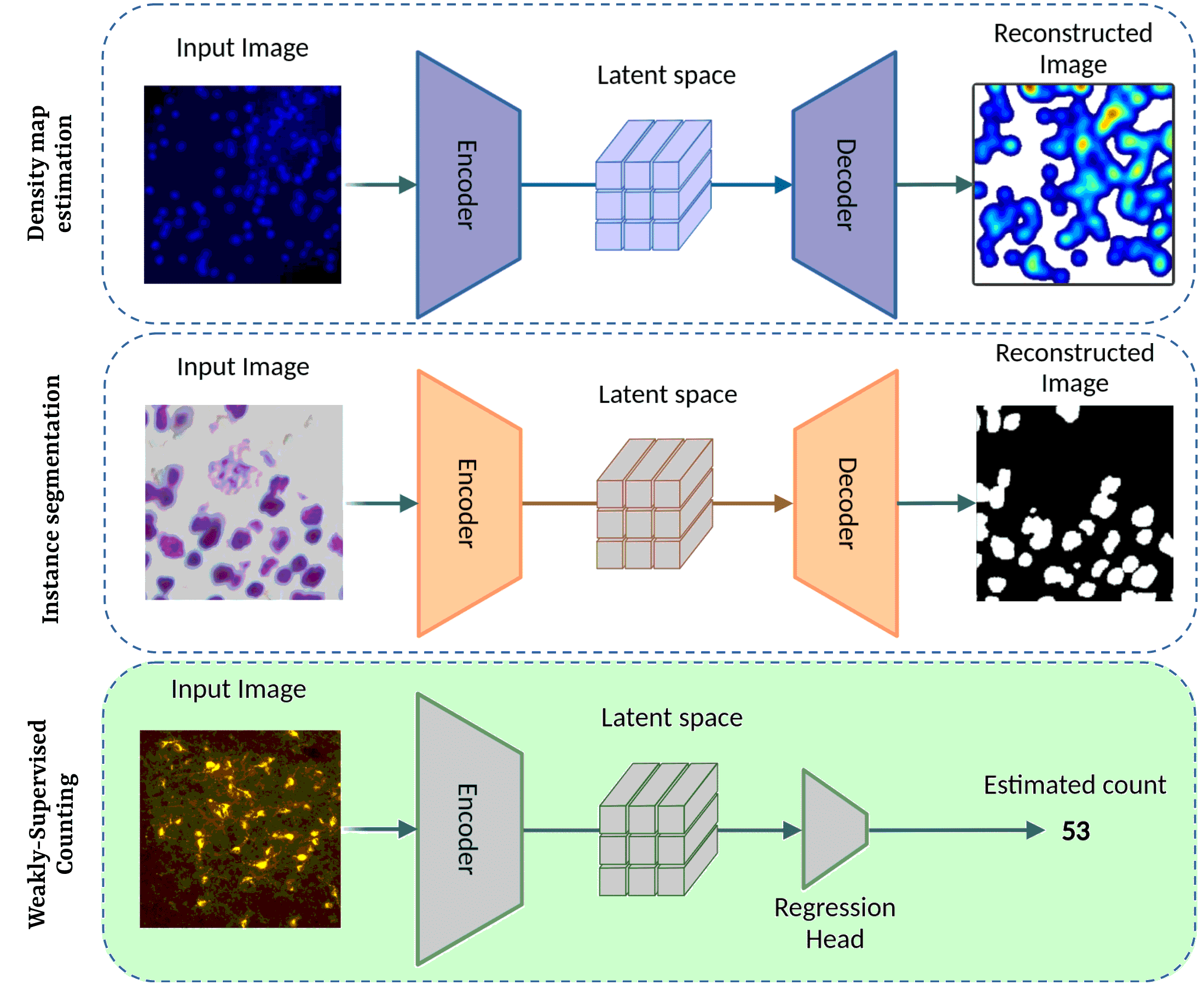}

    \caption{Comparison of methodologies in deep learning regarding instance counting. Most approaches tackle the solution by density map estimation or instance segmentation (top and middle). Weakly-supervised counting (bottom) regresses the total number of instances in the image, removing the need for spatial-aware ground truth.}
    \label{counting_methods_comparison}
    
\end{figure}

Although density estimation and instance segmentation offer benefits, they are not always feasible due to the lack of detailed spatial information. For example, microbial swab testing assesses surface cleanliness by providing a total bacterial count without precise localization \cite{Davidson1999EvaluationOT}. Spatial-aware datasets complicate enumeration by requiring detailed annotations and increasing computational burden \cite{Davidson1999EvaluationOT}. Instead, focusing on aggregate counts provides a simpler, faster, and equally effective solution.

To address this issue, weakly-supervised counting (WSC) is used as an approach that applies regression to images to predict the total number of instances without spatial information, as shown in Figure \ref{counting_methods_comparison}. CNNs are the most common architectural choice to solve this problem \cite{xue_WSC_cells_2016, ding_WSC_cells_2020}, but recent advances have shown the effectiveness of vision transformers (ViTs) for the same task, outperforming CNNs \cite{Liu_vit_density_2023, Liang_WSC_crowd_transf_2021}. This is the result of the inherent self-attention mechanisms of ViTs, which, in contrast to CNNs, outperform in capturing global image context and contextual dependencies, proving effective in image classification and segmentation \cite{dosovitskiy}.

The goal of this study is to highlight the use of ViTs in weakly-supervised counting and make use of its applicability in the task of microorganism enumeration as an effective solution. To achieve this, we conducted a comprehensive analysis of ViT-based backbone regression architectures. We compared them to the most popular benchmark architectures: CNNs, ResNet50, and ResNet101, by training them under the same strategy (no use of pre-trained weights nor fine-tuning to optimize results) to achieve the task of microorganism enumeration. We used four different microscopic-based datasets: the Fluorescent Neuronal Cells dataset \cite{cell_counting_yellow}, VGG-Cells dataset \cite{LTCO}, U2OS/HL60 Human Cancer Cells dataset \cite{CancerCells} and a self-made artificial fluorescent bacteria dataset, that is composed of high-resolution images. We created this dataset for the task of WSC in order to cover the gaps the former three datasets have in regards of dataset size, density sparsity through the images, and image resolution. 

Our experimental evaluation indicates that although traditional architectures such as ResNet achieve better performance, ViTs, especially CrossViT, can achieve comparable results to ResNet. CrossViT also performed exceptionally well on homogeneously distributed datasets, outperforming other ViT variants and CNNs in terms of computational efficiency. These results underscore the need to explore the use of ViTs, as further research holds the potential to achieve effective architectures for the task of weakly-supervised microorganism enumeration.

This study contributes to microorganism enumeration \cite{zhang_comprehensive_2022} by direct enumeration without the need for spatial information. We also analyze the use of vision transformers (ViTs) in regression counting. We evaluate popular and novel weakly-supervised counting methods, adapt them to microbial imaging, and evaluate ViTs in this context. Our analysis covers the capability of ViTs in a regression task, comparing different ViT approaches to identify the best one depending on the use case. By evaluating the performance of ViTs under the same training strategy as the other methods when configured for weakly-supervised counting, we provide insight into the limitations and potential of this architecture, thus contributing to the field of feature extraction using self-attention mechanisms \cite{khan_transformers_2022}. We also contribute to the study by developing an artificial fluorescent bacteria dataset designed for the task of weakly-supervised microorganism enumeration. An implementation of the method and the artificial dataset generation tool are available at \href{https://github.com/JavierUrenaPhDProjects/vits_for_WSC}{https://github.com/JavierUrenaPhDProjects/vits\_for\_WSC}.
\section{Related work}

\subsection{Machine Learning in microorganism enumeration}

Automating enumeration of microorganisms has been researched for decades, with traditional techniques like PCA, LDA or SVM and subsequently advancing to feature extraction with deep learning \cite{zhang_comprehensive_2022, Shabtai_ML_NN_1996}. Two research strategies have been developed: detection-based and regression-based enumeration. Detection-based methods refer to the enumeration of instances once they are located in the image, being image segmentation as the most accurate method of instance detection. \cite{Falk_UNet_2019, Hoorali_UNet_2020, Morelli_UNet_2021} extend the U-Net architecture \cite{Ronneberger_UNet_2015} in its ability to count cells and bacteria by discretizing them from background and subsequently enumerating them. Regression-based methods solve the task either through density map estimation or direct regression without spatial details. Convolutional neural networks (CNNs) for density estimation treats image pixels as real-valued feature vectors and constructs density functions over pixel grids, enabling object count estimation by integrating over specific image regions \cite{LTCO, xie_microscopy_CNN_density_2018, he_automatic_CNN_density_2019, he_deeply-supervised_CNN_density_2020}. The task is also accomplished with weak annotations, such as centroid position information of the instances, to estimate dense proximity maps \cite{Xie_CNN_density_2018, Zhang_CNN_density_2022}. 

But segmentation and density map regression in image analysis both rely on spatial information to identify instances, with segmentation using binary masks for pixel discretization \cite{Ronneberger_UNet_2015} and density regression assessing instance agglomeration \cite{LTCO}. Even weakly-supervised models sometimes require centroid annotations \cite{Zhang_CNN_density_2022}. However, in high-density scenarios common in microorganism analysis, detecting individuals becomes challenging. Moreover, when the goal is to simply count microorganisms globally, spatial details may be unnecessary. This necessitates datasets that bypass spatial data, focusing instead on correlating image features with total microorganism counts for efficiency \cite{Khan_WSC_cells_2016, xue_WSC_cells_2016, ding_WSC_cells_2020}. This is avoided through direct regression, or "\textit{weakly-supervised counting}" which is a form of supervised learning where the annotation of the images is kept to a minimum, such as providing only the overall global count \cite{ding_WSC_cells_2020}, or the patch-labeled count \cite{xue_WSC_cells_2016}. This approach emerges as an end-to-end application-oriented solution for microorganism enumeration.

\subsection{Weakly-supervised enumeration}

Weakly-supervised counting refer to the enumeration approach based on direct regression, and is tackled with machine learning when problems such as high instance density or occlusion arise, and also bypasses the hard detection problem and reduces labeling cost by requiring only the ground truth number of instances in the training images. This has been extensively studied in the use case of crowd counting, by the use of CNNs \cite{Liang_WSC_crowd_2023}, and microorganisms too \cite{ding_WSC_cells_2020, Lavitt_WSC_cells_2021}. \cite{Lavitt_WSC_cells_2021} for example implements an end-to-end WSC architecture by concatenating a ResNet with a CNN-based regressor that captures the global features of the entire microscopic image.

But CNNs convolution kernels fail to model global context information due to the limited receptive field, which is crucial when it comes to dense instance counting \cite{Liu_CNN_density_2019}, and do not establish interactions between image patches, which makes them unable to learn contrast features between the background and the elements to count. Vision transformers \cite{dosovitskiy} circumvent this issue with the self-attention mechanism that captures global dependencies and thus learn global context information. \textit{TransCrowd} \cite{Liang_WSC_crowd_transf_2021}, first uses ViT for weakly-supervised crowd counting as a backbone to extract information and concatenating a regression head, creating an end-to-end architecture. Others follow example. creating end-to-end architectures that achieve crowd counting with refined iterations \cite{Savner_WSC_crowd_transf_2022, Miao_WSC_crowd_transf_2023}.

\subsection{Object counting with transformer architectures}

As ViTs capture better global context information, most approaches use them as feature extraction backbone modules \cite{Li_vit_segmentation_survey_2023}. Images are segmented into fit-size patches, which are then processed through a linear embedding layer and sometimes summed with task-oriented tokens, which are then fed into the standard transformer encoder. The first use case where semantic segmentation was achieved using ViT instead of CNN as the backbone is \textit{SETR} \cite{Zheng_vit_segmentation_2021}, which achieved state-of-the-art results on the \textit{ADE20k} dataset.

For regression methods, transformers are considered as an effective solution to estimate density maps for crowd counting \cite{Sun_vit_density_2021, Tian_vit_density_2021, Liu_vit_density_2023} by using specialized patch tokens as a form is dense supervision \cite{Tian_vit_density_2021} or achieving multi-scale 2D feature maps from a pyramid ViT backbone, named \textit{CCTrans} \cite{Liu_vit_density_2023}. \textit{CounTR} \cite{Liu_vit_density_2023} leverages the instance counting task by creating a class-agnostic counting architecture by exploiting the attention mechanisms to explicitly capture the similarity between image patches or "exemplars".

Weakly-supervised counting has also been achieved using transformers \cite{Liang_WSC_crowd_transf_2021, Savner_WSC_crowd_transf_2022, Miao_WSC_crowd_transf_2023, Wang_WSC_crowd_transf_2022}. \textit{CCTwins} \cite{Dong_WSC_vit_2023}, uses a U-shaped architecture, featuring an adaptive Twins-SVT-L backbone to extract multi-level features, uses a multi-level count estimator to regress these features to a crowd number in a coarse-to-fine manner. \textit{Learn to Count Anything} \cite{Hobley_WSC_vit_2022} accomplishes class-agnostic instance counting like \textit{CounTR} \cite{Liu_vit_density_2023}, but without using exemplars or reference patches. Instead, it employs WSC with self-supervised knowledge distillation, where a teacher network processes global image slices and a student network processes smaller local slices.

Promising results have been achieved in the paradigm of instance counting using transformer-based architectures, but to date little to no research has been done in the field of microbiology. Research in this area can contribute to the development of more effective models for counting microorganisms.

\section{Methodology}
 
We subject a comparison of different architecture approaches covering three different categories: state-of-the-art approaches in the task of weakly-supervised counting, ViT-based backbones for regression architectures, and finally baseline traditional deep learning computer vision architectures commonly used for the task. The explanation can be followed in section \ref{ssec:proposed_architectures}.

The models were trained from scratch on four microorganism-based datasets consisting of neuronal cells, cancer cells, or bacteria. These datasets represent different types of use cases because they present different challenges, such as dataset size, density per image, or variability between images. The metrics used to compare the architectures are Mean Absolute Error (MAE) and Root Mean Squared Error (RMSE). Datasets and metrics are further explained in section \ref{ssec:experm}.  

\subsection{Architecture pipeline}\label{ssec:arch_pipeline}

\begin{figure}
    \centering

    \includegraphics[width=\linewidth]{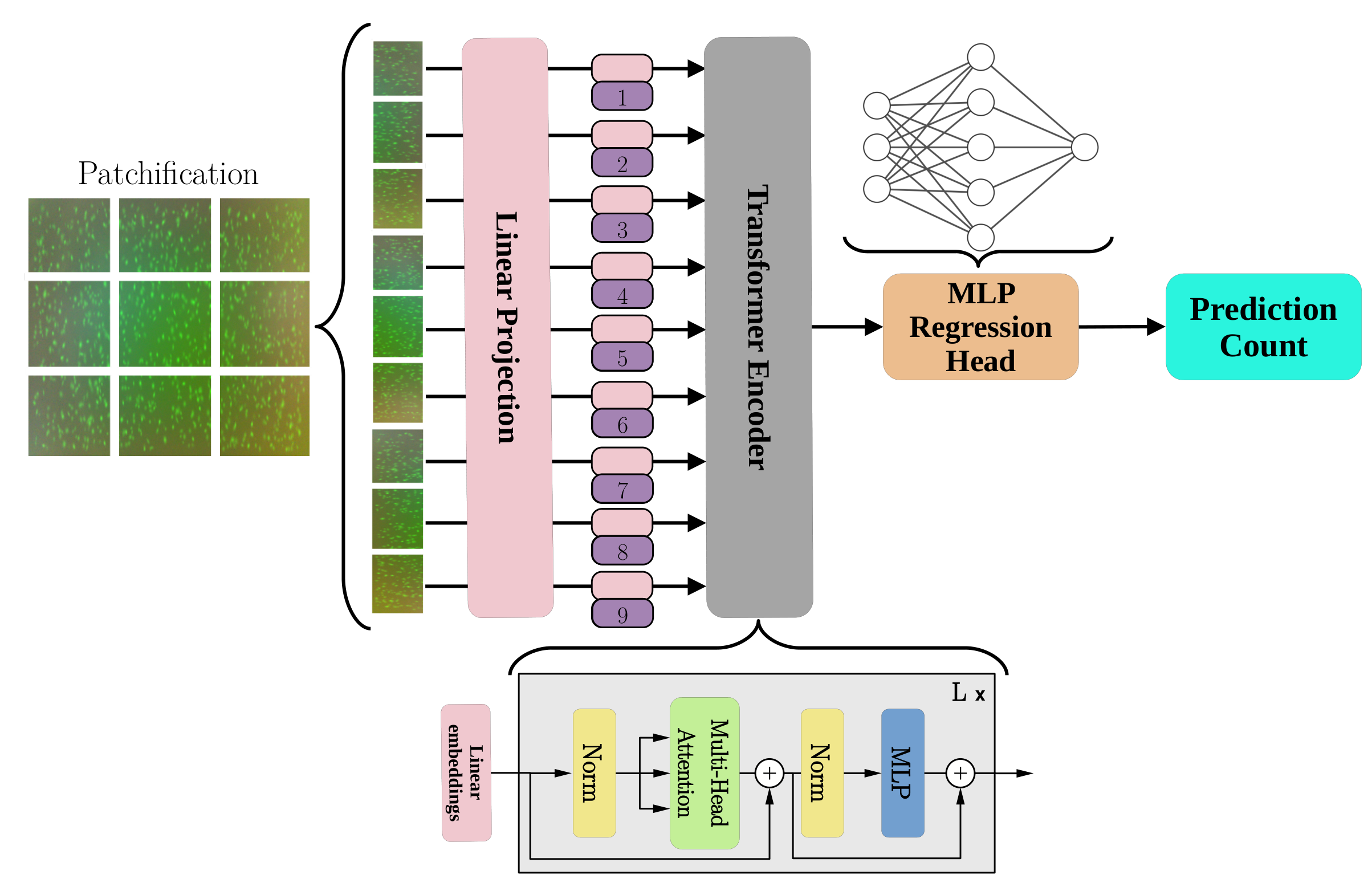}

    \caption{In ViT approaches for WSC, the ViT is used as a backbone, or feature extractor, and then concatenated with a regression head, or counter, which is usually an MLP that interprets the ViT embeddings into the number of instances in the image.}
    \label{fig:generic_arch}

\end{figure}

A common approach for WSC architectures is to concatenate two main parts: the backbone and the regression head (also called counter). The backbone is responsible for extracting the image features and placing the visual data into the latent space as feature embeddings. These are then sent to a regression head, which is used exclusively to predict the number of instances in the image. This general architectural approach, illustrated in the example of a ViT backbone in Figure \ref{fig:generic_arch}, is widely used in most studies investigating WSC \cite{Yang_WSC_crowd_2020, Liang_WSC_crowd_transf_2021, Lavitt_WSC_cells_2021, Savner_WSC_crowd_transf_2022, Wang_WSC_crowd_transf_2022}. Feature extraction is the most important part of the whole architecture, so this study focuses on exploring different backbones. The simplest regression head is a linear regressor, but the use of nonlinear regressors is preferable. In this study, it is implemented as a single layer fully connected network.

\subsection{Architecture backbones}\label{ssec:proposed_architectures}          

\subsubsection{SOTA architectures in WSC - TransCrowd}

TransCrowd \cite{Liang_WSC_crowd_transf_2021} is a pioneering ViT-based WSC model, featuring two implementations: TransCrowd-GAP and TransCrowd-Token. TransCrowd-GAP employs global average pooling on the transformer's output tokens, while TransCrowd-Token adds a learnable token for enumeration. This model shows significant improvements in crowd counting on datasets like ShanghaiTech, outperforming both weakly-supervised ($17.5\%$ MAE and $18.8\%$ MSE improvement over MATT \cite{Lei_WSC_crowd_2020}) and fully-supervised methods ($11.0/13.6$ MAE and $0.1/4.3$ MSE improvements compared to CRSNet and BL \cite{Li_CNN_density_2018, Ma_CNN_density_2019}).

\subsubsection{ViT backbones for WSC}
ViT research offers various backbones for feature extraction, chosen for their performance and proclaimed computational efficiency from novel architectural approaches. The first one being the vanilla ViT \cite{dosovitskiy}, which introduced the multi-head self-attention mechanism (MHSA) of the transformer as an encoder for image recognition, processing images as patch sequences. This method outperforms traditional CNNs in image classification benchmarks like ImageNet and CIFAR-100. 

A different ViT approach is the DeepViT \cite{zhou2021deepvit} which addresses "attention collapse", a problem with ViTs that make them plateau in performance when made deeper, by introducing "re-attention", a technique that regenerate attention maps with minimal computational cost, improving top-1 classification accuracy by 1.6\% on ImageNet with 32 transformer blocks.

An interesting approach to achieve great computational efficiency is CrossViT \cite{chen2021crossvit} which features a dual-branch transformer that processes different-sized patches with an efficient cross-attention mechanism that fuses these patches, reducing computational costs significantly. This model outperforms DeiT on ImageNet1K by 2\%, with minimal additional computational complexity and model size.

To achieve higher model complexity without compromising parameter and compute neutrality, Parallel ViT \cite{touvron2022parallelvit} proposes parallelizing the MHSA and feed-forward blocks by reorganizing the same blocks by pairs, resulting in the same number of parameters but wider and shallower, increasing the dimension of the embedding for better spatial feature separability.

Finally, to address the quadratic complexity of ViTs ($\mathcal{O}(w^2h^2)$), XCiT \cite{elnouby2021xcit} introduces cross-covariance attention (XCA), which applies self-attention across feature channels. This reduces the computational cost for high-resolution images while maintaining performance for WSC tasks common in bioinformatics. 

\subsubsection{Traditional computer vision architectures}

Two different common computer vision architectures are used as feature extractors. This will provide an unbiased baseline approach to achieve WSC more traditionally. The ResNet \cite{resnet} was chosen because it is commonly used as a feature extractor in both academia and industry services because its residual connections allow it to be deep while being computationally affordable. \cite{Lavitt_WSC_cells_2021} achieved WSC of cancer cells by implementing their version of ResNet called \textit{xResNet}. In this study, we implement ResNet50 and ResNet101 as competing backbones. Likewise, normal convolutional neural network backbones were also implemented, called CNN base, CNN medium, and CNN deep, each with different depths. These architectures are used to contrast the ResNet as computationally cheap architectures to achieve WSC.

\section{Experiments}
\begin{figure}
    \centering
    \includegraphics[width=0.9\linewidth]{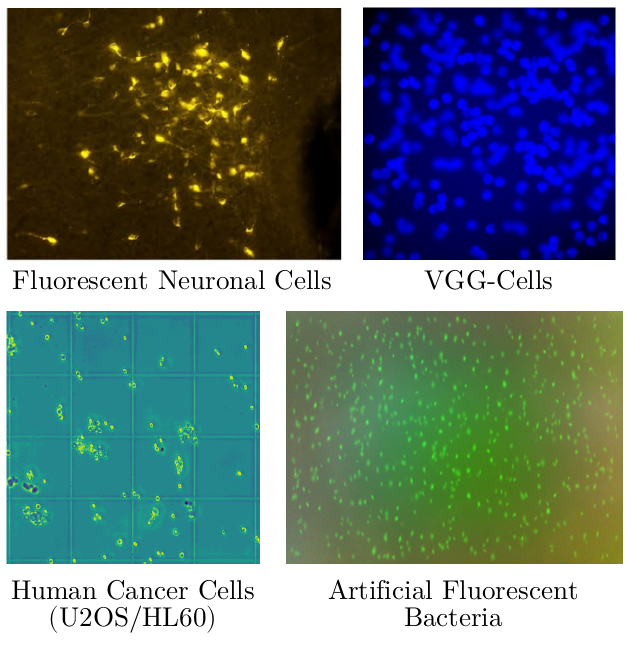}

    \caption{Random samples from the four datasets used in this study. Although the fluorescent neuronal cells and VGG-cells datasets were originally used for segmentation or density map estimation, their ground truths were analyzed and adapted for the task of WSC. We created the Artificial Fluorescent Bacteria to compensate the gaps the others datasets presented.}
    \label{fig:datasets_samples}

\end{figure}
\subsection{Implementation details}

The experimental framework was developed in Python 3.10, using the PyTorch library for model implementation and training, which supports CUDA GPU computation. The models are implemented from scratch in the case of the ResNets, and from the \textit{vit-pytorch}\footnote{\textit{Lucidrains} vit-pytorch Github page in the case of the ViT backbones: \url{https://github.com/lucidrains/vit-pytorch}} library was used, which faithfully implements the selected vision transformers and adapts them for WSC. The datasets go through a preprocessing stage of transformations for input normalization by the torchvision library: They are transformed into tensors, resized to a size of $384\times384$ pixels, normalized according to their corresponding mean and standard deviation characteristics, and finally processed as 32-bit floating point values for computational ease. Then, depending on the type of architecture, the images are tokenized (for transformers) at different patch sizes, depending on the configuration described in each architecture's respective paper: $16\times16$ for implementations of TransCrowd, XCiT, Parallel ViT, or DeepViT, and $32\times32$ for ViT. CrossViT uses both patch sizes since it works at multi-granularity. In ResNets, the images are entered as a whole. The architectural implementation of each type of model is defined by the hyperparameter configuration in its own paper. The table \ref{tab:arch_param} summarizes the architectural properties of each chosen model variant.

\begin{table*}[t] 
\centering
\caption{Characteristics of the considered architectures. The variants are defined by their original papers. Depth refers to the number of blocks or layers, heads to the size of the MHSA, dimension to the size of the embedding dimensionality, and MLP dim to the convolutional output in the case of CNNs and ResNet, and to the MLP head inner dimensionality for the ViT cases.}
\label{tab:arch_param}
\small
\begin{tabular*}{\textwidth}{@{\extracolsep{\fill}}lcccccc@{}}
\toprule
\text{Model name} & \text{Variant} &
\text{Depth} & 
\text{Heads} &  
\text{Dimension} &
\text{MLP Dim.} & 
\text{\begin{tabular}[c]{@{}c@{}}Number of\\ parameters (\num{e6})\end{tabular}} \\ 
\midrule
CNN      & Base/Medium/Deep & 1 / 2 / 3 & - & - & 16 / 64 / 256 & 0.59 / 0.61 / 0.96\\ 
ResNet  & 50/101 & 16 / 33 & - & - & 2048 & 23.53 / 42.54\\
ViT      & Vanilla & 12 & 12 & 768 & 3072 & 87.50\\
XCiT   & S24 & 24 & 8 & 384 & 1536 & 49.82 \\ 
CrossViT  & Ti & 4 & 3 &  96 / 192 & 384 / 768 & 3.07\\ 
Parallel ViT  & Ti & 12 & 3 & 192 & 192 & 5.50 \\ 
Deep ViT     & S & 16 & 12 & 396 & 1188 & 34.91 \\
TransCrowd-G  & Vanilla & 12 & 12 & 768 & 3072 & 90.39 \\
TransCrowd-T  & Vanilla & 12 & 12 & 768 & 3072 & 86.86 \\
\bottomrule
\end{tabular*}
\end{table*}

Training uses a heuristic approach of end-to-end training with no pre-trained weights, so models are trained from scratch and with nearly identical training hyperparameters. The batch size is determined by the computational load of the architecture, using 128 for smaller architectures (CNNs), 64 for Vanilla ViT, CrossViT, TransCrowd, and ResNet, and 32 for Parallel ViT, Deep ViT, and XCiT. The training learning rate is initially defined as $10^{-4}$ and then dynamically configured with a scheduler that reduces its value by a factor of \num{2e-4} each time the validation loss plateaus a patience window of 5 epochs. The transformer architectures count with a "warm-up" period of 5000 steps before training, where the optimization starts with a minimum learning rate that is gradually increased to a predefined maximum in a predefined number of steps. This is done because experiments show that the gradients in the final layers of the transform tend to be large, causing an exploding gradient effect if the learning rate is too high \cite{onlayernormalization}.
The fixed number of epochs for all models is 400, although training is interrupted when the validation plateaus after a patience of 20 epochs. Experiments were run with different randomization seeds and the training was performed on an NVIDIA GeForce RTX 4090 GPU.
\begin{table*}[t]
\centering
\caption{Characteristics of the datasets used. The number of images refers to the size of the training and validation sets, the resolution refers to their original datasets, the ground truth is how they were originally labeled, and the count statistics show the total number of instances in all images of the dataset, and the minimum, average, and maximum number of instances per image.}
\label{tab:datasets_comparison}
\small
\renewcommand{\arraystretch}{1.0}
\begin{tabular*}{\textwidth}{@{\extracolsep{\fill}}lcccccccc@{}}
\toprule
 &  &  &  &  & \multicolumn{4}{c}{Count statistics} \\
\cmidrule(lr){6-9}
Dataset & 
\multicolumn{1}{c}{\begin{tabular}[c]{@{}c@{}}Number of\\ images\end{tabular}} & 
\multicolumn{1}{c}{\begin{tabular}[c]{@{}c@{}}Data\\ augmentation\end{tabular}} & 
\multicolumn{1}{c}{\begin{tabular}[c]{@{}c@{}}Resolution\end{tabular}} & 
\multicolumn{1}{c}{\begin{tabular}[c]{@{}c@{}}Type of\\ ground truth\end{tabular}} & 
\text{Total number} & 
\text{Min} & 
\text{Ave} & 
\text{Max} \\ 
\midrule
Fluorescent Neuronal Cells & $10000$ & yes & $1600\times1200$ & Binary mask & $18475$ & $0$ & $2$ & $17$ \\
VGG-Cells & $19294$ & yes & $256\times256$ & Point annotation & $2480424$ & $1$ & $128$ & $317$ \\
Human Cancer Cells & $1463$ & yes & $700\times700$ & Global count & $201058$ & $0$ & $138$ & $410$ \\
Artificial Fluorescent Bacteria & $12000$ & no & $3280\times2464$ & Point annotation & $10963874$ & $0$ & $913$ & $1855$ \\ 
\bottomrule
\end{tabular*}
\end{table*}

\subsection{Datasets and metrics}\label{ssec:experm}

Four different datasets related to the fluorescent microbial paradigm have been used for comparison, with different characteristics of color, size, shape and quantity: The \textit{Fluorescent Neuronal Cells} \cite{cell_counting_yellow}, \textit{VGG-Cells} \cite{LTCO}  and \textit{Human Cancer Cells} \cite{CancerCells} datasets are freely available for research purposes. These datasets allow us to analyze each model under different use cases, all of which are summarized in the table \ref{tab:datasets_comparison}.

\paragraph{Fluorescent Neuronal Cells \cite{cell_counting_yellow}} A collection of 283 high-resolution images (1600x1200 pixels) of neurons from mouse brains. The original ground truth represent segmentation masks of the different neurons. We used the watershed algorithm to discretize the different cells in the masks to obtain the number of cells per image. After patching the images and augmenting the data, the total size of the training and validation sets is 10000 images, where for each image the number of brain cells present in the image was assigned as a label. This dataset challenges the models to regress information from complex images with small numbers. It's available from the AMSActa repository\footnote{AMSActa Repository: \url{http://amsacta.unibo.it/id/eprint/6706/}}. 

\paragraph{VGG Cells \cite{LTCO}} Created using the system in \cite{bluecelldatasetgeneration}, this dataset counts on image ground truths that place pixel indications of the centroid of each cell. The original dataset has 200 images of 256x256 pixel resolution, along with their respective ground truths of the same resolution. After applying data augmentation, 19294 images were used for training and validation. To obtain the total number of cells, we counted the pixel centroids from the ground truth of each image. This dataset is challenging due to high density of instances per image and high occlusion. It is available from the University of Oxfords Visual Geometry Group publication page\footnote{Oxford University Visual Geometry Group: \url{https://www.robots.ox.ac.uk/~vgg/research/counting/index_org.html}}.

\paragraph{Human Cancer Cells \cite{CancerCells}} Contains microscope images of a human osteosarcoma cell line (U2OS) and a human leukemia cell line (HL-60). The dataset was originally prepared for the cell counting task and contains 165 labeled images, of which 133 are used for training. After applying data augmentation, we reach a total number of 1463 training and validation images. This dataset challenges the models to achieve good convergence from a small amount of data. It's available in the Zenodo data repository \footnote{Microscope images of human cancer cell lines - Zenodo: \url{https://zenodo.org/records/4428844}}.

\paragraph{Artificial Fluorescent Bacteria Dataset} In collaboration with the Department of Chemistry at the University of Innsbruck, Austria, a series of fluorescent microbial microscope images has been created. These images consist of a series of bacteria of the type \textit{Bacillus Subtilis} that are suspended and captured by a digital microscope. The main problem is that these images are not labeled, unlike the public datasets that came with labeled ground truths. With this motivation, we created an artificial dataset of fluorescent bacteria where we could have our own labeled data for the WSC task and also compensate the gaps that the other datasets present (have a large dataset, large and low density per image, and homogeneous sparsity in all images). To create such a dataset, we collected a series of samples of the microscopic images background without any bacteria present, and proceeded to randomly place "fluorescent bacteria" in it. As the usual shape of a bacterium, these artificial bacteria are formed according to a Gaussian ellipse of random size, axis and rotation determine the pixel intensity in the RGB channels to color the fluorescence, and then placed in a random pixel coordinate in the image, so that they appear realistic and natural. Thus, a total of 12000 images were generated for a data set containing images with bacteria ranging from 0 to nearly 1900 bacteria per image.  

\begin{table*}
\centering
\caption{Average mean absolute error, root mean square error, and floating point operations per second. The best result is shown in bold and the second best in underlined. Separated are the three types of architectures chosen in the experiments: traditional methods, state-of-the-art, and ViT backbone implementations.}
\label{tab:models_performance}
\begin{tabularx}{\textwidth}{
  >{\hsize=1.3\hsize\raggedright\arraybackslash}X
  *{8}{>{\hsize=0.94\hsize\centering\arraybackslash}X}
  >{\hsize=1.2\hsize\centering\arraybackslash}X} 
\toprule
& \multicolumn{2}{c}{\begin{tabular}[c]{@{}c@{}}Fluorescent\\ Neuronal Cells\end{tabular}} & \multicolumn{2}{c}{VGG-Cells} & \multicolumn{2}{c}{\begin{tabular}[c]{@{}c@{}}Human\\ Cancer Cells\end{tabular}}  & \multicolumn{2}{c}{\begin{tabular}[c]{@{}c@{}}Fluorescent\\ Artificial Bacteria\end{tabular}} & {}\\
   Architectures &  MAE &  RMSE &  MAE & RMSE &  MAE & RMSE &  MAE & RMSE & FLOPS (\num{e8})\\
\midrule
CNN Base       & 2.123          & 3.769          & 6.212          & 8.072         & 59.584        & 86.751            & 43.282          & 54.534            & 0.53 \\
CNN Medium     & 1.828          & 3.346          & 4.991          & 7.129         & 40.130        & 65.224            & 34.307          & 46.354            & 7.86 \\
CNN Deep       & 1.566          & \textbf{2.861} & 1.827          & 2.891         & 52.507        & 79.794            & 24.987          & 31.082            & 56.48 \\
ResNet50       & 1.508          & 3.229          & \textbf{1.225} & \underline{1.815} & \textbf{27.206} & \textbf{42.341} & \underline{22.030} & \underline{29.028}  & 120.19 \\
ResNet101      & \textbf{1.400} & \underline{2.871} & \underline{1.311} & \textbf{1.699} & \underline{30.497} & \underline{45.526} & 23.313         & 30.726            & 229.58 \\
\midrule
TransCrowd-G   & 2.011          & 3.859          & 9.321          & 13.421         & 47.627        & 68.871            & 31.997         & 44.456            & 554.86 \\
TransCrowd-T   & 1.497          & 2.999          & 2.585          & 3.401         & 58.152        & 79.791            & 22.563          & 32.220            & 554.64 \\
\midrule
CrossViT       & 1.634          & 3.274          & 3.012          & 3.794         & 43.356        & 69.466            & \textbf{20.011} & \textbf{27.677}   & 50.91 \\
DeepViT        & \underline{1.464} & 3.025          & 37.713         & 57.980        & 71.171        & 98.258            & 106.402         & 146.692           & 293.68 \\
XCiT           & 1.654          & 3.243          & 3.139          & 4.100         & 49.163        & 69.083            & 27.158          & 36.490            & 265.03 \\
Parallel ViT   & 1.519          & 3.167          & 2.400          & 2.975         & 44.972        & 63.012            & 23.736          & 31.244            & 62.42 \\
Vanilla ViT    & 1.541          & 3.257          & 1.886          & 2.714         & 57.886        & 84.347            & 26.055          & 35.760            & 128.39 \\
\bottomrule

\end{tabularx}

\end{table*}

For performance evaluation, we chose mean absolute error (MAE) and root mean square error (RMSE) as the main metrics. MAE, which is the average of the difference between the true and predicted values, gives a straightforward answer to how accurate the model is in estimating the count, since in many cases it is critical to have the same level of predictability in both high and low density image scenarios. RMSE gives greater weight to larger errors, which is helpful when measuring models that have larger discrepancies than smaller ones. RMSE helps to understand how the model behaves when large quantities appear in the image. We also measure the average FLOPS (floating point operations per second) during inference to analyze efficiency.

\section{Results}

To evaluate different approaches to the task of weakly-supervised enumeration of microorganisms, we tested several deep learning architectures (CNNs and ViTs) on four different microorganism-based datasets. The results of our evaluation are presented in the following sections.

\subsection{Performance evaluation through datasets}
The results of the performance evaluation, as presented in Table \ref{tab:models_performance}, demonstrate the relative efficacy of different architectural approaches across a range of datasets. For fluorescent neurons with low instance density, deeper convolutional neural networks (CNNs) and residual networks (ResNets) demonstrate enhanced performance, with ResNets achieving the highest accuracy. Additionally, ViT-based architectures demonstrate satisfactory performance, with DeepViT exhibiting the most favorable outcomes. In the VGG-cells dataset, which has a high instance density per image, ResNets demonstrated the highest level of performance. However, the vanilla ViT variant exhibited a notable reduction in the performance gap, emerging as the top-performing ViT variant in this high-density scenario. In the case of the U2OS/HL60 human cancer dataset, which is distinguished by a lower density and a smaller data size, all models demonstrated suboptimal performance. This is likely attributable to the limited number of training images. Despite the fact that ViTs require larger datasets than ResNets in order to excel, they were unable to close the performance gap with the latter. Nevertheless, Parallel ViT, CrossViT, and XCiT outperformed the deep CNN. In the homogeneous Artificial Bacteria dataset, CrossViT surpassed traditional architectures, likely due to its dual-branch multi-scale feature representation. This demonstrates the potential of ViT-based architectures when optimal data conditions are met.

\subsection{Performance evaluation of TransCrowd}
The TransCrowd models \cite{Liang_WSC_crowd_transf_2021} exhibited varying performance across the datasets. The TransCrowd-Token model demonstrated the most favorable performance on average, outperforming the TransCrowd-GAP model in the fluorescent neuronal cells, VGG-cells, and artificial bacteria datasets. In our experiments, we found that the learnable regression token from TransCrowd-Token achieved better results than TransCrowd-GAP in the fluorescent neuronal cells, VGG-cells, and artificial bacteria datasets. Specifically, we observed an improvement in MAE/RMSE of 25.55\%/22.28\%, 72.26\%/74.66\%, and 29.48\%/27.52\%, respectively. In the Human Cancer Cells dataset, TransCrowd-GAP outperforms TransCrowd-Token by 18.10\% and 13.69\%, respectively, indicating that global average pooling of the transformer output can facilitate better generalization in smaller datasets. TransCrowd-Token demonstrated a level of competitiveness, achieving superior outcomes to those of the average ViT backbone approaches in studies. In fact, they achieved comparable results to ResNets in the Fluorescent Neuronal Cells and Fluorescent Artificial Bacteria datasets. However, they exhibited the slowest performance in terms of inference time, which negatively impacted their overall efficiency.

\subsection{Evaluation of computational efficiency}

CNN models demonstrate optimal computational efficiency, achieving the lowest FLOPs in all architectures, although this does not align with their overall performance. In contrast, ResNet models demonstrate an optimal balance of performance across all four datasets, while maintaining a moderate computational expense. ViT based architectures, on the other hand, are resource-intensive, as even with a relatively small number of learnable parameters, they remain highly complex, resulting in a slower inference time. However, this is not the case for the chosen architecture of CrossViT, which achieves \num{50.91e8} FLOPS, much faster than any other ViT architecture or ResNet, and even faster than the deep CNN, despite having three times the number of learnable parameters (\num{0.96e6} versus \num{3.07e6}). This can be explained by the CrossViT architecture's balance of complexity while still achieving fine-grained patch sizes for the transformer encoder, which makes it more computationally efficient. This results in a discrepancy with the Parallel ViT approach, which was originally designed to be lightweight without compromising performance, and performed with \num{62.42e8} FLOPS, which is 18.45\% slower than the CrossViT.
\section{Discussion}
This study advances our understanding of using deep learning for microorganism enumeration \cite{zhang_comprehensive_2022} and explores the capabilities of transformer architectures in computer vision \cite{khan_transformers_2022}. It evaluates different architectures under uniform training conditions to objectively assess how vision transformers compare to current methods in weakly-supervised microorganism enumeration.

Our experimental evaluation shows that training vision transformers from scratch for weakly-supervised counting is challenging because these models are typically designed for tasks such as image classification or segmentation. Therefore, most studies use pre-trained weights and special fine-tuning for ViT-based approaches. This is manifested in the performance of TransCrowd \cite{Liang_WSC_crowd_transf_2021}, which, after pre-training and fine-tuning in its original use case, showed state-of-the-art performance, with TransCrowd-GAP outperforming TransCrowd-Token, contrary to the results of our study. Nevertheless, our experiments showed that ViT-based models provide comparable results, motivating further research to find more effective and scalable ViT-based regression models.

From a practical perspective, traditional ResNets are effective for microorganism enumeration, as they are efficient and can converge even with small datasets. This efficiency is crucial for routine laboratory applications where large labeled datasets are impractical or costly. However, ViTs have demonstrated superior performance regarding WSC in other fields under specific configurations \cite{Miao_WSC_crowd_transf_2023, Wang_WSC_crowd_transf_2022, Savner_WSC_crowd_transf_2022, Liang_WSC_crowd_transf_2021}, suggesting that with proper care, they could surpass benchmarks in microorganism enumeration as well.

Future work would include exploring generalization through cross-validation across different datasets, and investigating the performance of vision transformers (ViTs) when optimally pre-trained and fine-tuned, together in comparison with a broader list of baseline architectures like DenseNet or GoogLeNet, and analyzing their applicability by using real-world bacteria/cells samples, in contrast with laboratory, environment controlled data. As current ViT research is primarily focused on improving tasks such as image classification, weakly-supervised enumeration remains underexplored, hence, future work could focus on improving the ViTs for this specific task.
\section{Conclusion}

We have analyzed different vision transformer-based architectures, including state-of-the-art models like TransCrowd and together with common deep learning computer vision approaches for the task of weakly-supervised enumeration of microorganisms. We have evaluated these architectures on heuristic training through four different microorganism-based datasets to analyze the capabilities of vision transformers in a regression task. We show that although current standard approaches such as residual networks outperform ViTs, the latter is relevant for providing feature extraction to be used for weakly-supervised counting of microorganisms through regression. We further show that microorganism enumeration can be solved from a weakly-supervised counting perspective, providing insight into the potential of using ViT for more adaptable and scalable approaches.

\section{Acknowledgments}
This paper is part of the research and development project DesDet in collaboration with the department of Analytical Chemistry and Radiochemistry, Hollu Systemhygiene GmbH and Planlicht GmbH \& Co KG. This project is funded by Standortagentur Tirol.
\bibliographystyle{ieeetr}          

\bibliography{references}

\end{document}